\documentclass[10pt,twocolumn,letterpaper]{article}

\usepackage[pagenumbers]{cvpr} % To force page numbers, e.g. for an arXiv version

\usepackage[dvipsnames]{xcolor}

\usepackage{algorithm2e}
\usepackage{float}
\usepackage{array}
\usepackage{csquotes}
\usepackage{colortbl}
\definecolor{cvprblue}{rgb}{0.21,0.49,0.74}
\usepackage[pagebackref,breaklinks,colorlinks,citecolor=cvprblue]{hyperref}
\usepackage{multirow}

\title{MCTrack: A Unified 3D Multi-Object Tracking Framework for Autonomous Driving}

\author{
Xiyang Wang\textsuperscript{1}\thanks{Equal contribution.} \and 
Shouzheng Qi\textsuperscript{2}\footnotemark[1] \and 
Jieyou Zhao\textsuperscript{3}\footnotemark[1] \and 
Hangning Zhou\textsuperscript{4}\thanks{Corresponding author.  Email: zhouhangning@megvii.com} \and 
Siyu Zhang\textsuperscript{1} \and 
Guoan Wang\textsuperscript{1} \and 
Kai Tu\textsuperscript{1} \and 
Songlin Guo\textsuperscript{1} \and 
Jianbo Zhao \textsuperscript{5} \and 
Jian Li\textsuperscript{2} \and 
Mu Yang\textsuperscript{4} \and 
 \\[0.5em] 
\textsuperscript{1} Mach Drive   \,  \textsuperscript{2} National University of Defense Technology \\ 
\, \, \, \textsuperscript{3} Sichuan University   \, \textsuperscript{4} MEGVII Technology   \, \textsuperscript{5} University of Science and Technology of China
}

\begin{document}
\maketitle
\begin{abstract}
This paper introduces MCTrack, a new 3D multi-object tracking method that achieves state-of-the-art (SOTA) performance across KITTI, nuScenes, and Waymo datasets. Addressing the gap in existing tracking paradigms, which often perform well on specific datasets but lack generalizability, MCTrack offers a unified solution. Additionally, we have standardized the format of perceptual results across various datasets, termed BaseVersion, facilitating researchers in the field of multi-object tracking (MOT) to concentrate on the core algorithmic development without the undue burden of data preprocessing. Finally, recognizing the limitations of current evaluation metrics, we propose a novel set that assesses motion information output, such as velocity and acceleration, crucial for downstream tasks. The source
codes of the proposed method are available at this link:
\href{https://github.com/megvii-research/MCTrack}{https://github.com/megvii-research/MCTrack}
\end{abstract}    
\section{Introduction}
\label{sec:intro}

\begin{figure}
    \centering
    \includegraphics[width=1.0\linewidth]{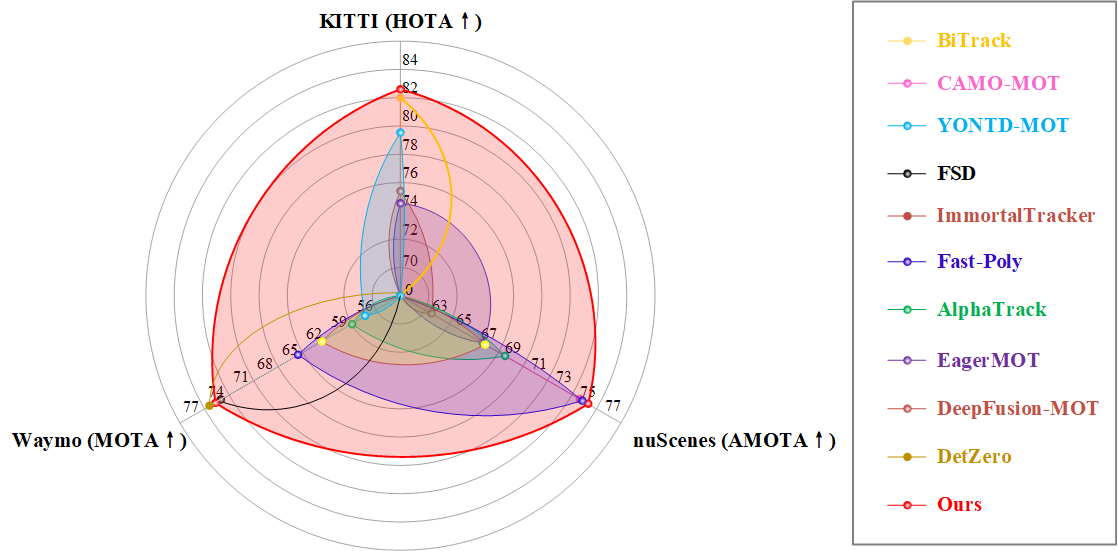}
    \caption{The comparison of the proposed method with SOTA methods across different datasets. For the first time, we have achieved SOTA performance on all three datasets.
}
    \label{fig:radar}
\end{figure}

3D multi-object tracking plays an essential role in the field of autonomous driving, as it serves as a bridge between perception and planning tasks. The tracking results directly affect the performance of trajectory prediction, which in turn influences the planning and control of the ego vehicle. Currently, common tracking paradigms include tracking-by-detection (TBD) \cite{weng20203d,wojke2017simple,wang2022deepfusionmot}, tracking-by-attention (TBA) \cite{chu2023transmot,sun2020transtrack,zeng2022motr}, and joint detection and tracking (JDT) \cite{wang2023you,bergmann2019tracking}.  Generally, the TBD paradigm approach tends to outperform the TBA and JDT paradigm methods in both performance and computational resource efficiency. Commonly used datasets include KITTI \cite{geiger2012we}, Waymo \cite{sun2020scalability}, and nuScenes \cite{caesar2020nuscenes}, which exhibit significant differences in terms of collection scenarios, regions, weather, and time. Furthermore, the difficulty and format of different datasets vary considerably. Researchers often need to write multiple preprocessing programs to adapt to different datasets. The variability across datasets typically results in these methods attaining SOTA performance solely within the confines of a particular dataset, with less impressive results observed on alternate datasets \cite{li2023poly,wang2023camo,huang2024bitrack,li2024fast}, as shown in  Fig.\ref{fig:radar}. For instance, DetZero \cite{ma2023detzero} achieved SOTA performance on the Waymo dataset but was not tested on other datasets. Fast-Poly \cite{li2024fast} achieved SOTA performance on the nuScenes dataset but had mediocre performance on the Waymo dataset. Similarly, DeepFusion \cite{wang2022deepfusionmot} performed well on the KITTI dataset but exhibited average performance on the nuScenes dataset. Furthermore, in terms of performance evaluation, existing metrics such as CLEAR \cite{bernardin2008evaluating}, AMOTA \cite{weng20203d}, HOTA \cite{luiten2021hota}, IDF1 \cite{ristani2016performance}, etc., mainly judge whether the trajectory is correctly connected. They fall short, however, in evaluating the precision of subsequent motion information—key information such as velocity, acceleration, and angular velocity—which is crucial for fulfilling the requirements of downstream prediction and planning tasks \cite{li2023poly,wang2022deepfusionmot,wang2023camo}.

Addressing the noted challenges, we first introduced the BaseVersion format to standardize perception results (i.e., detections)  across different datasets. This unified format greatly aids researchers by allowing them to focus on advancing MOT algorithms, unencumbered by dataset-specific discrepancies.

Secondly, this paper proposes a unified multi-object tracking framework called MCTrack. To our best knowledge, our method is the first to achieve SOTA performance across the three most popular tracking datasets: KITTI, nuScenes, and Waymo. Specifically, it ranks first in both the KITTI and nuScenes datasets, and second in the Waymo dataset. It is worth noting that the detector used for the first-place ranking in Waymo dataset is significantly superior to the detector we employed. Moreover, this method is designed from the perspective of practical engineering applications, with the proposed modules addressing real-world issues. For example, our two-stage matching strategy involves the first stage, which performs most of the trajectory matching on the bird's-eye view (BEV) plane. However, for camera-based perception results, matching on the BEV plane can encounter challenges due to the instability of depth information, which can be as inaccurate as 10 meters in practical engineering scenarios. To address this, trajectories that fail to match in the BEV plane are projected onto the image plane for secondary matching. This process effectively avoids issues such ID-Switch (IDSW) and Fragmentation (Frag) caused by inaccurate depth information, further improving the accuracy and reliability of tracking.

Finally, this paper introduces a set of metrics for evaluating the motion information output by MOT systems, including speed, acceleration, and angular velocity. We hope that researchers will not only focus on the correct linking of trajectories but also consider how to accurately provide the motion information needed for downstream prediction and planning after correct matching, such as speed and acceleration.
\section{Related Work}
\label{sec:RelatedWork}

\subsection{Datasets}
Multi-object tracking can be categorized based on spatial dimensions into 2D tracking on the image plane and 3D tracking in the real world. Common datasets for 2D tracking methods include MOT17 \cite{milan2016mot16}, MOT20 \cite{dendorfer2020mot20}, DanceTrack \cite{sun2022dancetrack}, etc., which typically calculate 2D IoU or appearance feature similarity on the image plane for matching \cite{aharon2022bot,cao2023observation,maggiolino2023deep}. However, due to the lack of three-dimensional information of objects in the real world, these methods are not suitable for applications like autonomous driving. 3D tracking methods often utilize datasets such as KITTI \cite{geiger2012we}, nuScenes \cite{caesar2020nuscenes}, Waymo \cite{sun2020scalability}, which provide abundant sensor information to capture the three-dimensional information of objects in the real world. Regrettably, there is a significant format difference among these three datasets, and researchers often need to perform various preprocessing steps to adapt their pipeline, especially for TBD methods, where different detection formats pose a considerable challenge to researchers. To address this issue, this paper standardizes the format of perceptual results (detections) from the three datasets, allowing researchers to focus better on the study of tracking algorithms.

\subsection{MOT Paradigm}
Common paradigms in multi-object tracking currently include Tracking-by-Detection (TBD) \cite{wang2022deepfusionmot,zhang2303bytetrackv2}, Joint Detection and Tracking (JDT) \cite{bergmann2019tracking,wang2023you}, Tracking-by-Attention (TBA) \cite{chu2023transmot,sun2020transtrack}, and Referring Multi-Object Tracking (RMOT) \cite{wu2023referring,du2024ikun}. JDT, TBA, and RMOT paradigms typically rely on image feature information, requiring GPU resources for processing. However, for the computing power available in current autonomous vehicles, supporting the GPU resources needed for MOT tasks is impractical. Moreover, the performance of these paradigms is often not as effective as the TBD approach. Therefore, this study focuses on TBD-based tracking methods, aiming to design a unified 3D multi-object tracking framework that accommodates the computational constraints of autonomous vehicles.

\subsection{Data Association}
In current 2D and 3D multi-object tracking methods, cost functions such as IoU, GIoU \cite{rezatofighi2019generalized}, DIoU \cite{zheng10faster}, Euclidean distance, and appearance similarity are commonly used \cite{cao2023observation,aharon2022bot}. Some of these cost functions only consider the similarity between two bounding boxes, while others focus solely on the distance between the centers of the boxes. None of them can ensure good performance for each category in every dataset. The Ro\_GDIoU proposed in this paper, which takes into account both shape similarity and center distance, effectively addresses these issues. Moreover, in terms of matching strategy, most methods adopt a two-stage approach: the first stage uses a set of thresholds for matching, and the second stage relaxes these thresholds for another round of matching. Although this method offers certain improvements, it can still fail when there are significant fluctuations in the perceived depth. Therefore, this paper introduces a secondary matching strategy based on the BEV plane and the Range View (RV) plane, which solves this problem effectively by matching from different perspectives.

\subsection{MOT Evaluation Metrics}
The earliest multi-object tracking evaluation metric, CLEAR, was proposed in reference \cite{bernardin2008evaluating}, including metrics such as MOTA and MOTP. Subsequently, improvements based on CLEAR have led to the development of IDF1 \cite{ristani2016performance}, HOTA \cite{luiten2021hota}, AMOTA \cite{weng20203d}, and so on. These metrics primarily assess the correctness of trajectory connections, that is, whether trajectories are continuous and consistent, and whether there are breaks or ID switches. However, they do not take into account the motion information that must be output after a trajectory is correctly connected in a multi-object tracking task, such as velocity, acceleration, and angular velocity. This motion information is crucial for downstream tasks like trajectory prediction and planning. In light of this, this paper introduces a new set of evaluation metrics that focus on the motion information output by MOT tasks, which we refer to as motion metrics. We encourage researchers in the MOT field to focus not only on the accurate association of trajectories but also on the quality and suitability of the trajectory outputs to meet the requirements of downstream tasks.

\section{MCTrack}
\label{sec:MCTrack}

\begin{figure*}[h]
    \centering
    \includegraphics[width=1\linewidth]{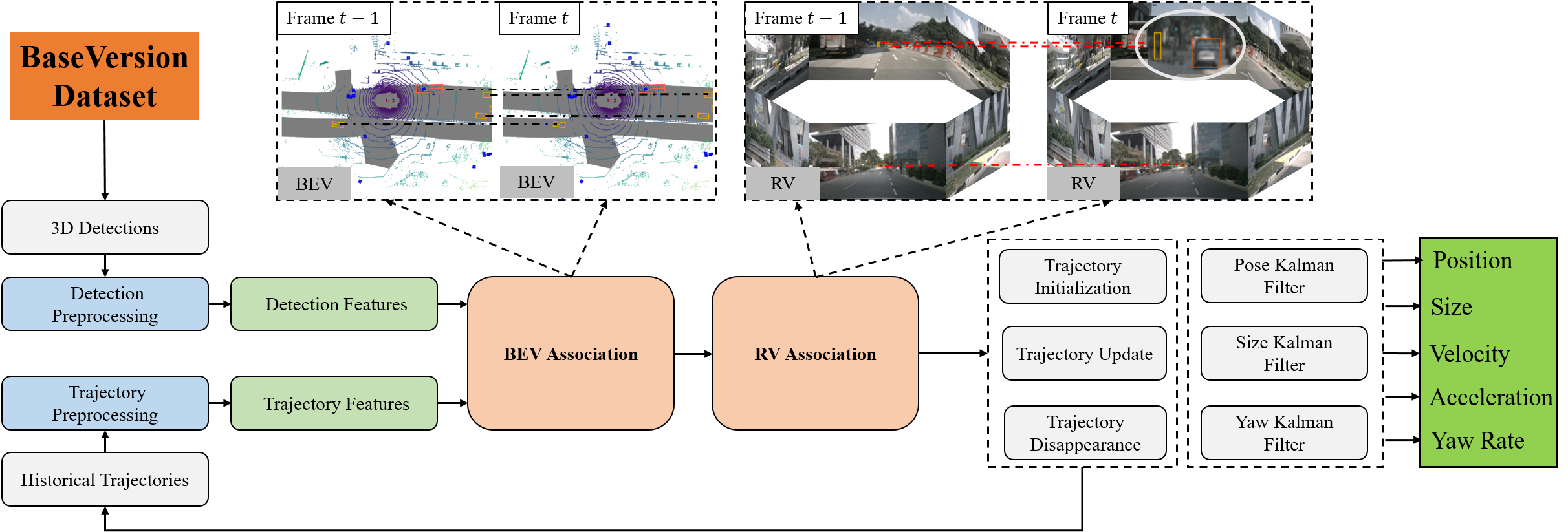}
    \caption{Overview of our unified 3D MOT framework MCTrack. Our input involves converting datasets such as KITTI, nuScenes, and Waymo into a unified format known as BaseVersion. The entire pipeline operates within the world coordinate system. Initially, we project 3D point coordinates from the world coordinate system onto the BEV plane for the primary matching phase. Subsequently, unmatched trajectory boxes and detection boxes are projected onto the image plane for secondary matching. Finally, the state of the trajectories is updated, along with the Kalman filter. Our output includes motion information such as position, velocity, and acceleration, which are essential for downstream tasks like prediction and  planning.
}
    \vspace{-10pt}
    \label{fig:PPL}
\end{figure*}

We present MCTrack, a streamlined, efficient, and unified 3D multi-object tracking method designed for autonomous driving. The overall framework is illustrated in Fig.\ref{fig:PPL}, and detailed descriptions of each component are provided below.

% 3.2
\subsection{Data Preprocessing}

\begin{figure}[H]
    \centering
    \includegraphics[width=1.0\linewidth]{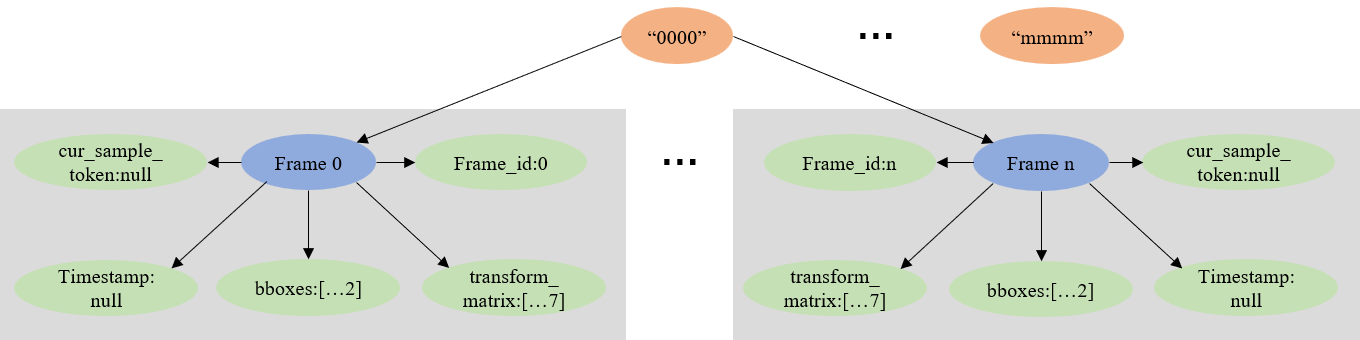}
    \caption{BaseVersion data format overview.}
    \label{fig:baseversion}
\end{figure}

To validate the performance of a unified pipeline (PPL) across different datasets and to facilitate its use by researchers, we standardized the format of detection data from various datasets, referring to it as the BaseVersion format. This format encapsulates the position of obstacles within the global coordinate system, organized by scene ID, frame sequence, and other pertinent parameters. As depicted in Figure \ref{fig:baseversion}, the structure includes a comprehensive scene index with all associated frames. Each frame is detailed with frame number, timestamp, unique token, detection boxes, transformation matrix, and additional relevant data.

 For each detection box, we archive details such as \enquote{detection score,} \enquote{category,} \enquote{global\_xyz,} \enquote{lwh,} \enquote{global\_orientation} 
  (expressed as a quaternion), \enquote{global\_yaw} in radians, \enquote{global\_velocity,} \enquote{global\_acceleration.} For more detailed explanations, please refer to our code repository.

% 3.2
\subsection{MCTrack Pipeline}

\subsubsection{Kalman Filter}
Currently, most 3D MOT methods \cite{zhang2303bytetrackv2,wu20213d,li2023poly} incorporate position, size, heading, and score into the Kalman filter modeling, esulting in a state vector ${S=\left \{ x,y,z,l,w,h,\theta ,score,v_x, v_y,v_z \right \}}$  that can have up to 11 dimensions, represented using a unified motion equation, such as constant velocity or constant acceleration models. It is important to note that in this paper, $\theta$ specifically denotes the heading angle. However, this modeling approach has the following issues: Firstly, different state variables may have varying units (e.g., meters, degrees) and magnitudes (e.g., position might be in the meter range, while scores could range from 0 to 1), which can lead to numerical stability problems. Secondly, some state variables exhibit nonlinear relationships (such as the periodic nature of angles), while others are linear (such as dimensions), making it challenging to represent them with a unified motion equation. Furthermore, combining all state variables into a single model increases the dimensionality of the state vector, thereby increasing computational complexity. This may reduce the efficiency of the filter, particularly in real-time applications. Therefore, we decouple the position, size, and heading angle, applying different Kalman filters to each component.

For position, we only need to model the center point $x, y$ in the BEV  plane using a constant acceleration motion model. The state and observation vectors are defined as follows:
\begin{equation}
S_\textup{p} = \left \{ x, y, v_x, v_y, a_x, a_y \right \} \quad
M_\textup{p} = \left \{ x, y, v_x, v_y \right \}
\end{equation}

For size, we only use the length and width ${l, w}$ with a constant velocity motion model. The state vector and observation vector are represented as follows:
\begin{equation}
S_\textup{s}=\left \{ l,w,v_l,v_w \right \} \quad
M_\textup{s}=\left \{ l,w\right \} 
\end{equation}
\noindent It should be noted that, theoretically, the size of the same object should remain constant. However, due to potential errors in the perception process, we rely on filters to ensure the stability and continuity of the size.

For heading angle, we  use the constant velocity motion model. The state vector and observation vector are represented as follows:
\begin{equation}
S_\mathrm{\theta}=\left \{ \theta_p, \theta_v, \omega_p, \omega_v \right \} \quad
M_\mathrm{\theta}=\left \{ \theta_p, \theta_v \right \} 
\end{equation}
\noindent Here, ${\theta_p}$ denotes the heading angle provided by perception, while ${\theta_v}$ represents the heading angle calculated from velocity, that is ${\theta_v=\arctan ({v_y}/{v_x}) }$. 

\subsubsection{Cost Function}

\begin{figure}
    \centering
    \includegraphics[width=1.0\linewidth]{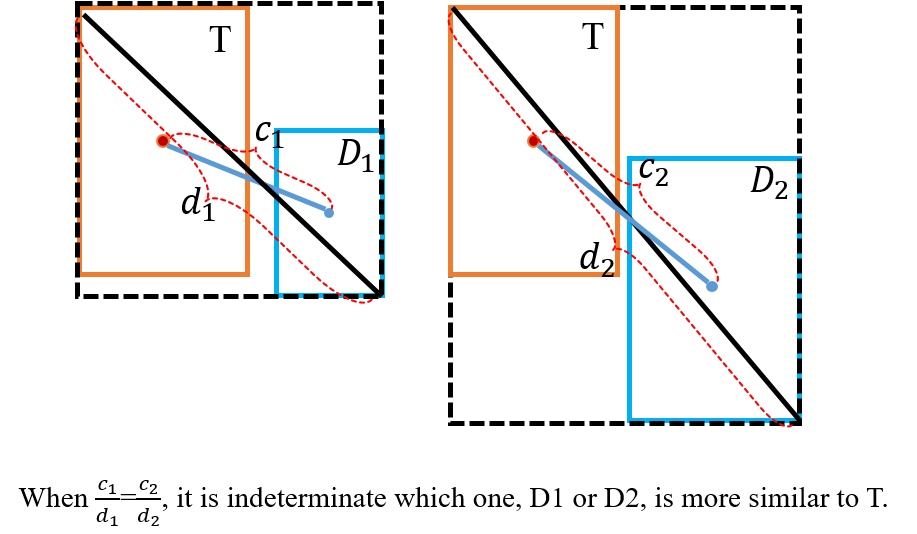}
    \caption{The problem existing in the tracking field with ${DIoU}$.}
    \label{fig:diou}
\end{figure}

\begin{figure}
    \centering
    \includegraphics[width=1.0\linewidth]{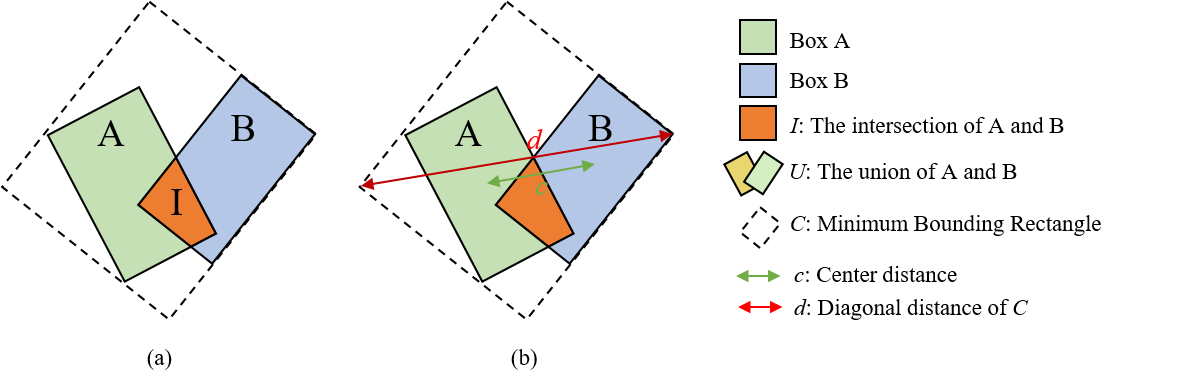}
    \caption{Schematic of ${Ro\_GDIoU}$ calculation.}
    \label{fig:rogdiou}
\end{figure}

 As indicated in reference \cite{zheng2020distance}, GIoU fails to distinguish the relative positional relationship when two boxes are contained within one another, effectively reducing to IoU. Similarly, for DIoU, problems also exist, as shown in Fig.\ref{fig:diou}. When the IoU of two boxes is 0 and the center distances are equal, it is also difficult to determine the similarity between the two boxes.  Our extensive experiments reveal that using only Euclidean distance or IoU and its variants as the cost metric is inadequate for capturing similarity across all categories. However, combining distance and IoU yields better results. To address these limitations, we propose ${Ro\_GDIoU}$, an IoU variant based on the BEV plane that incorporates the heading angle of the detection box by integrating ${GIoU}$ and ${DIoU}$. Fig.\ref{fig:rogdiou} shows a schematic of the ${Ro\_GDIoU}$ calculation, and the corresponding pseudocode is provided in Algorithm \ref{alg:RoGDIOU}.

% ---------- ROGDIOU伪代码 --------------------------
\SetKwComment{Comment}{/* }{ */}
\RestyleAlgo{ruled}
\begin{algorithm}
\caption{Pseudo-code of $ \mathit {Ro\_GDIoU} $}
\label{alg:RoGDIOU}
\KwIn{Detection bounding box $B^{\mathrm{d}} = (x^{\mathrm{d}}, y^{\mathrm{d}}, z^{\mathrm{d}}, 
    l^{\mathrm{d}}, w^{\mathrm{d}}, h^{\mathrm{d}}, \theta^{\mathrm{d}})$ and Trajectory bounding box ${B}^{\mathrm{t}} = (x^{\mathrm{t}}, y^{\mathrm{t}}, z^{\mathrm{t}}, l^{\mathrm{t}}, w^{\mathrm{t}}, h^{\mathrm{t}}, \theta^{\mathrm{t}})$}
\KwOut{$\mathrm {Ro\_GDIoU}$}
    ${B}^{\mathrm{d}}_{\mathrm{bev}}, {B}^{\mathrm{t}}_{\mathrm{bev}} = \mathcal{F}_{\mathrm{global\longrightarrow bev}}(B^{\mathrm{d}}, {B}^{\mathrm{t}})$\;
    Calculate the area of intersection $ \mathcal{I} = \mathcal{F}_\mathrm{inter}({B}^{\mathrm{d}}_{\mathrm{bev}}, {B}^{\mathrm{t}}_{\mathrm{bev}})$\;
    Calculate the area of union $ \mathcal{U} = \mathcal{F}_\mathrm{union}({B}^{\mathrm{d}}_{\mathrm{bev}}, {B}^{\mathrm{t}}_{\mathrm{bev}})$\;
    Calculate the minimum enclosing rectangle $ \mathcal{C} = \mathcal{F}_\mathrm{rect}({B}^{\mathrm{d}}_{\mathrm{bev}}, {B}^{\mathrm{t}}_{\mathrm{bev}})$\;
    Calculate the Euclidean distance between the center points of the two boxes $ {c} = \mathcal{F}_\mathrm{dist}({B}^{\mathrm{d}}_{\mathrm{bev}}, {B}^{\mathrm{t}}_{\mathrm{bev}})$\;
    Calculate the diagonal distance of of the minimum enclosing rectangle  $ {d} = \mathcal{F}_\mathrm{dist}({B}^{\mathrm{d}}_{\mathrm{bev}}, {B}^{\mathrm{t}}_{\mathrm{bev}})$\;

    $ \mathit {Ro\_IoU} = \frac{\mathcal{I}}{\mathcal{U}}$\;
    $ \mathit {Ro\_GDIoU} = \mathit{Ro\_IoU} - \omega_{1} \cdot \frac{\mathcal{C} - \mathcal{U}}{\mathcal{C}} - \omega_{2} \cdot \frac{{c}^{2}}{{d}^{2}}$\
\end{algorithm}
% ====================== END =======================

Where ${\omega_{1}}$ and ${\omega_{2}}$ represent the weights for ${IoU}$ and Euclidean distance respectively, and ${\omega_{1}}$ + ${\omega_{2}}$ = 2. When two bounding boxes perfectly match, ${Ro\_IoU}$ = 1,  ${\frac{C - U}{C}}$=${\frac{c^{2}}{d^{2}}}$=0,  which means ${{Ro\_GDIoU} = 1}$. When two boxes are far away, ${Ro\_IoU} = 0$, ${\frac{C - U}{C}}$=${\frac{c^{2}}{d^{2}}} {\to}$ 1, which means ${{Ro\_GDIoU}}$ = -2.

When calculating the ${Ro\_GDIoU}$ between the detection box and the trajectory box, we combine forward trajectory predictions using Kalman filtering with backward predictions based on detections. Assuming the detection box at time $\tau$ is represented as $ D_{\tau} =\left \{ {D_{\tau}^{i}} \right \}_{i=1}^{N} \subset \mathbb{R} ^{D^{X}\times 1 } $, with ${D_{\tau}^{i}}=\left \{ {x^{\mathrm{d}}_{\tau},y^{\mathrm{d}}_{\tau},z^{\mathrm{d}}_{\tau},l^{\mathrm{d}}_{\tau},w^{\mathrm{d}}_{\tau},h^{\mathrm{d}}_{\tau},\theta^{\mathrm{d}}_{\tau},v_{\mathrm{x},\tau}^{\mathrm{d}}, v_{\mathrm{y},\tau}^{\mathrm{d}}} \right \} $. and the trajectory at time ${\tau-1}$ is represented as  $ T_{\tau-1} =\left \{ {T_{\tau-1}^{j}} \right \}_{j=1}^{M} \subset \mathbb{R} ^{T^{X}\times 1 } $, with ${T_{\tau-1}^{j}}=\left \{ {x^{\mathrm{t}}_{\tau-1},y^{\mathrm{t}}_{\tau-1},z^{\mathrm{t}}_{\tau-1},l^{\mathrm{t}}_{\tau-1},w^{\mathrm{t}}_{\tau-1},h^{\mathrm{t}}_{\tau-1},\theta^{\mathrm{t}}_{\tau-1},v^{\mathrm{t}}_{\mathrm{x},\tau-1}, v^{\mathrm{t}}_{\mathrm{y},\tau-1}} \right \} $.
The forward prediction can be computed as follows:
\begin{equation}
  T_{\tau}^{j}=\mathcal{F} (x^{\mathrm{t}}_{\tau-1}, y^{\mathrm{t}}_{\tau-1}, v^{\mathrm{t}}_{\mathrm{x},\tau-1}, v^{\mathrm{t}}_{\mathrm{y},\tau-1}, \Delta \tau).
  \label{eq:back_predict}
\end{equation}
\noindent where, ${\mathcal{F(\cdot )}}$ represents the motion equation, and in this case, we adopt the constant velocity model. The variable ${\Delta \tau}$ represents the time difference between the current frame and the previous frame.

The backward prediction can be computed as follows:
\begin{equation}
  D_{\tau-1}^{i}=\mathcal{F}^{-1} (x^{\mathrm{d}}_{\tau}, y^{\mathrm{d}}_{\tau}, v^{\mathrm{d}}_{\mathrm{x},\tau}, v^{\mathrm{d}}_{\mathrm{y},\tau}, -\Delta \tau).
  \label{eq:back_predict}
\end{equation}

Ultimately, the cost function between the detection box and the trajectory box is computed by the following formula:
\begin{equation}
\mathcal{L}_{cost} =\alpha \cdot \mathcal{C}(D_{\tau}, T_{\tau})+ (1-\alpha)\cdot \mathcal{C}(D_{\tau-1}, T_{\tau-1}).
\end{equation}
where, $\alpha \in [0, 1]$, and $C$ represents ${Ro\_GDIoU}$.

\subsubsection{Two-Stage Matching}
Similar to most methods \cite{zhang2303bytetrackv2,li2023poly}, our pipeline also utilizes a two-stage matching process, with the specific flow shown in Pseudocode \ref{alg:Matching}. However, the key difference is that our two-stage matching is performed from different perspectives, rather than by adjusting thresholds within the same perspective.

The calculations for \( T_{\text{bev}} \) and \( D_{\text{bev}} \) are illustrated in equation \ref{eq:cal_bev}. For the calculation of SDIoU, please consult the approach detailed in \cite{wang2022strongfusionmot}. We define the coordinate information of the detection or trajectory box as \( X = [x, y, z, l, w, h, \theta] \). According to equation \ref{eq:cal_bev}, we can determine the corresponding 8 corners, denoted as \( C = [P_0, P_1, P_2, P_3, P_4, P_5, P_6, P_7] \). Among these corners, we select the points with indices [2, 3, 7, 6] to represent the 4 points on the BEV plane.

\begin{equation}
\label{eq:cal_bev}
C = R \cdot  P + T.
\end{equation}

\begin{equation}
\renewcommand{\arraystretch}{1.2} % 调整行间距
\setlength{\arraycolsep}{2pt} % 调整列间距
P=
\begin{bmatrix}
\frac{l}{2}  \! & \frac{l}{2}  & \frac{l}{2}  & \frac{l}{2}  & -\frac{l}{2}  & -\frac{l}{2}  & -\frac{l}{2} \! & -\frac{l}{2} \\
\frac{w}{2}  & -\frac{w}{2} & -\frac{w}{2} & \frac{w}{2} & \frac{w}{2} & -\frac{w}{2} & -\frac{w}{2} & \frac{w}{2}\\
\frac{h}{2}  & \frac{h}{2} & -\frac{h}{2} & -\frac{h}{2} & \frac{h}{2} & \frac{h}{2} & -\frac{h}{2} & -\frac{h}{2}
\end{bmatrix}.
\end{equation}

\begin{equation}
R = \begin{bmatrix}
\cos(\theta) & -\sin(\theta) & 0 \\
\sin(\theta) & \cos(\theta) & 0 \\
0 & 0 & 1
\end{bmatrix},
\hspace{1em}
T = \begin{bmatrix}
x \\
y \\
z
\end{bmatrix}.
\end{equation}

% ---------- 两阶段匹配的伪代码 --------------------------
\SetKwComment{Comment}{/* }{ */}
\RestyleAlgo{ruled}
\begin{algorithm}
\caption{Pseudo-code of Two-stage Matching}
\label{alg:Matching}
\KwIn{Trajectory boxes $T$ at time $\tau-1$, detection boxes $D$ at time $\tau$}
\KwOut{Matching indices $ \mathcal{M}$ }
/* \textit{First matching: BEV Plane} */\;
$T_\mathrm{bev}, D_\mathrm{bev} = \mathcal{F}_{\mathrm{3d}\longrightarrow \mathrm{bev}}(T, D)$\;
Compute Cost $\mathcal{L}_\mathrm{bev}  = \text{Ro\_GDIoU}(D_\mathrm{bev}, T_\mathrm{bev})$.\;
Matching pairs $ \mathcal{M}_\mathrm{bev} = \text{Hungarian}(\mathcal{L}_\mathrm{bev}, \text{threshold}_\mathrm{bev})$\;
/* \textit{Second matching: RV Plane} */\;
\For{$d_\mathrm{bev}$ \textnormal{in} $D_\mathrm{bev}$}{
  \If{$d_\mathrm{bev}$ \textnormal{not in} $M_\mathrm{bev}[:, 0]$}{
  $d_\mathrm{bev} \longrightarrow D^\mathrm{res}$ \;
  }
}
\For{$t_\mathrm{bev}$ \textnormal{in}  $T_\mathrm{bev}$}{
  \If{$t_\mathrm{bev}$ \textnormal{not in} $M_\mathrm{bev}[:, 1]$}{
  $t_\mathrm{bev} \longrightarrow T^\mathrm{res}$ \;
  }
}
$D_\mathrm{rv}^\mathrm{res}, T_\mathrm{rv}^\mathrm{res} = \mathcal{F}_{\mathrm{3d}\longrightarrow \mathrm{rv}}(D^\mathrm{res}, T^\mathrm{res})$\;
Compute Cost $\mathcal{L}_{\mathrm{rv}}  = \text{SDIoU}(D_\mathrm{rv}^\mathrm{res}, T_\mathrm{rv}^\mathrm{res})$.\;
Matching pairs $\mathcal{M}_\mathrm{rv} = \text{Greedy}(\mathcal{L}_\mathrm{rv}, \text{threshold}_\mathrm{rv})$\;
Obtain the final matching pairs $\mathcal{M} = \mathcal{M}_\mathrm{bev}  \cup  \mathcal{M}_\mathrm{rv}$\;
\end{algorithm}
% ====================== END =======================

\section{New MOT Evaluation Metrics}
\label{sec:METRICS}

% 4.1
\subsection{Static Metrics}

Traditional MOT evaluation primarily relies on metrics such as CLEAR \cite{bernardin2008evaluating}, AMOTA  \cite{weng20203d}, HOTA \cite{luiten2021hota}, and IDF1 \cite{ristani2016performance}. These metrics focus on assessing the correctness and consistency of trajectory connections. In this paper, we refer to these metrics as static metrics. However, static metrics do not consider the motion information of trajectories after they are connected, such as speed, acceleration, and angular velocity. In fields like autonomous driving and robotics, accurate motion information is crucial for downstream prediction, planning, and control tasks. Therefore, relying solely on static metrics may not fully reflect the actual performance and application value of a tracking system. 

Introducing motion metrics into MOT evaluation to assess the motion characteristics and accuracy of trajectories becomes particularly important. This not only provides a more comprehensive evaluation of the tracking system's performance but also enhances its practical application in autonomous driving and robotics, ensuring that the system meets real-world requirements and performs effectively in complex environments.

% 4.2
\subsection{Motion Metrics}

To address the issue that current MOT evaluation metrics do not adequately consider motion attributes, we propose a series of new motion metrics, including Velocity Angle Error (VAE), Velocity Norm Error (VNE), Velocity Angle Inverse Error (VAIE), Velocity Inversion Ratio (VIR),  Velocity Smoothness Error (VSE) and Velocity Delay Error (VDE). These motion metrics aim to comprehensively assess the performance of tracking systems in handling motion features, covering the accuracy and stability of motion information such as speed, angle, and velocity smoothness.

% VAE
VAE represents the error between the velocity angle obtained from tracking cooperation and the ground truth angle, calculated as:

\begin{equation}
    \mathrm{VAE} = (\theta_{{gt}} - \theta_{{d}} + \pi) \bmod 2\pi - \pi.
  \label{eq:VAE}
\end{equation}
% 公式10改成(gt - dt + pi) mod 2*pi - pi
where $\theta_{\text{gt}}$ denotes the angle calculated from the target speed, and $\theta_{\text{d}}$ denotes the angle calculated from the tracking speed, with both angles ranging from $0^\circ$ to $360^\circ$. Given the discontinuity of angles, a $1^\circ$ difference from $359^\circ$ effectively corresponds to a $2^\circ$ separation. 

% VAIE
VAIE quantifies the angle error when the velocity angle error surpasses a predefined threshold of  $\frac{1}{2}\pi $. Breaching this threshold typically indicates that the tracking system's estimation of the target's velocity direction is directly opposite to the actual direction.

\begin{equation}
  \mathrm{VAIE}=\left|\theta_{\mathrm{gt}}-\theta_{\mathrm{d}}\right|, \quad \text { if }\left|\theta_{\mathrm{gt}}-\theta_{\mathrm{d}}\right|>\frac{1}{2}\pi .
  \label{eq:VAIE}
\end{equation}

% VIR
The corresponding VIR stands for velocity inverse ratio, which represents the proportion of velocity angle errors that exceed the threshold.

\begin{equation}
     \mathrm{ VIR }=\frac{\sum_{i=1}^{N}\pi_{i}}{N} , \pi_{i}=\left\{\begin{array}{ll}
    1, & \text { if VAIE exists } \\
    0, & \text { otherwise }.
    \end{array}\right.
  \label{eq:VIR}
\end{equation}
where $N$ represents the sequence length of the trajectory.

% VNE
VNE represents the error between the magnitude of velocity obtained by the tracking system and the true magnitude of velocity, calculated as:

\begin{equation}
   \mathrm{VNE}=\left| {V}_{\mathrm{gt}}-{V}_{\mathrm{d}}\right|.
  \label{eq:VNE}
\end{equation}
where ${V}_{\mathrm{gt}}$ and ${V}_{\mathrm{d}}$ represent the actual and predicted velocity magnitudes, respectively. 

% VSE
VSE represents the smoothness error of the velocity obtained from the filter. The smoothed velocity is calculated using the Savitzky-Golay (SG) \cite{schafer2011savitzky} filter.

\begin{equation}
\begin{aligned}
{V}_{\mathrm{d}}^{SG}={SG}\left({V}_{\mathrm{d}}, w, p\right) ,
\end{aligned}
\label{eq:VSE}
\end{equation}

\begin{equation}
\begin{aligned}
\mathrm{VSE}=\| {V}_{\mathrm{d}}-{V}_{\mathrm{d}}^{SG}) \|.
\end{aligned}
\label{eq:VSE}
\end{equation}
where $w$ and $p$ refer to the window size and polynomial order of the filter, respectively. ${V}_{\mathrm{d}}^{SG}$ represents the velocity value after being smoothed by the ${SG}$ filter. A smaller $\mathrm{VSE}$ value indicates that the original velocity curve is smoother.

% VDE
VDE represents the time delay of the velocity signal obtained by the tracking system relative to the true velocity signal. It is calculated by finding the offset within a given time window, which minimizes the sum of the mean and standard deviation of the difference between the true velocity and the velocity obtained by the tracking system.

First, we use a peak detection algorithm to identify the set $v_{gt}^{p}$ of local maxima in the velocity ground truth sequence.

\begin{equation}
\begin{aligned}
v_{\mathrm{gt}}^{p} = \mathcal{F} (v_{\mathrm{gt}}),
\end{aligned}
\label{eq:VDE}
\end{equation}
Here, $\mathcal{F}(\cdot)$ denotes the peak detection function, where the peak points must satisfy the condition $v_{\text{gt}}[t - 1] < v_{\text{gt}}[t] > v_{\text{gt}}[t + 1]$. Subsequently, we calculate the difference between the ground truth velocity and the tracking velocity within a given time window.

\begin{equation}
\begin{aligned}
{V}_{\mathrm{gt}}^{w} = {V}_{\mathrm{gt}}[t-{w/2} : t+{w/2}],
\end{aligned}
\label{eq:VDE}
\end{equation}

\begin{equation}
\begin{gathered}
{V}_{\mathrm{d},{\tau}}^{w} =  {V}_{\mathrm{d}}[t -{w/2} + {\tau} :  t+{w/2} + {\tau}],  \\
\tau \in \left[0, {n}\right]
\end{gathered}
\label{eq:VDE}
\end{equation}

\begin{equation}
\begin{aligned}
\Delta {V}_{\tau}^{w} = \{ \left | {v}_{\mathrm{gt}}^{i} - {v}_{\mathrm{d},{\tau}}^{i} \right | \mid ({v}_{\mathrm{gt}}^{i} \in {V}_{\mathrm{gt}}^{w}, {v}_{\mathrm{d},{\tau}}^{i} \in {V}_{\mathrm{d},{\tau}}^{w}) \},
\end{aligned}
\label{eq:VDE}
\end{equation}
where $t$ represents the time corresponding to the peak point, $w$ represents the window length, $\tau$ indicates the shift length applied to the velocity window from the tracking system, and $\Delta {V}_{\tau}^{w}$ represents the set of differences between the true velocity and the tracking velocity. Next, we calculate the mean and standard deviation of set $\Delta {V}_{\tau}^{w}$.

\begin{equation}
\begin{aligned}
M = \{ \mu_0, \mu_1, ..., \mu_{n} |  \mu_{\tau} = \frac{1}{w} \sum_{j=1}^{w} \Delta v_{\tau}^{j} \},
\end{aligned}
\label{eq:VDE}
\end{equation}

\begin{equation}
\begin{aligned}
\Sigma = \{ \sigma_0, \sigma_1, ..., \sigma_{n} | \sigma_{\tau} =\sqrt{\frac{1}{w} \sum_{j=1}^{w}\left(\Delta v_{\tau}^{j}-\mu_{\tau} \right)^{2}} \},
\end{aligned}
\label{eq:VDE}
\end{equation}

Finally, the time offset $\tau$ corresponding to the minimum sum of the mean and standard deviation is the VDE. 

\begin{equation}
\begin{aligned}
\mathrm{VDE}= \tau = \arg\min_{\tau \in [0, n]} \left( M + \Sigma \right).
\end{aligned}
\label{eq:VDE}
\end{equation}

\noindent where $\tau$ is the timestamp corresponding to the velocity vector and $n$ is the considered time window. It is important to note that for the ground truth of a trajectory, there can be multiple peak points in the time series. The above calculation method only addresses the lag of a single peak point. If there are multiple peak points, the average will be taken to represent the lag of the entire trajectory.

\begin{figure}[h]
    \centering
    \includegraphics[width=1.0\linewidth]{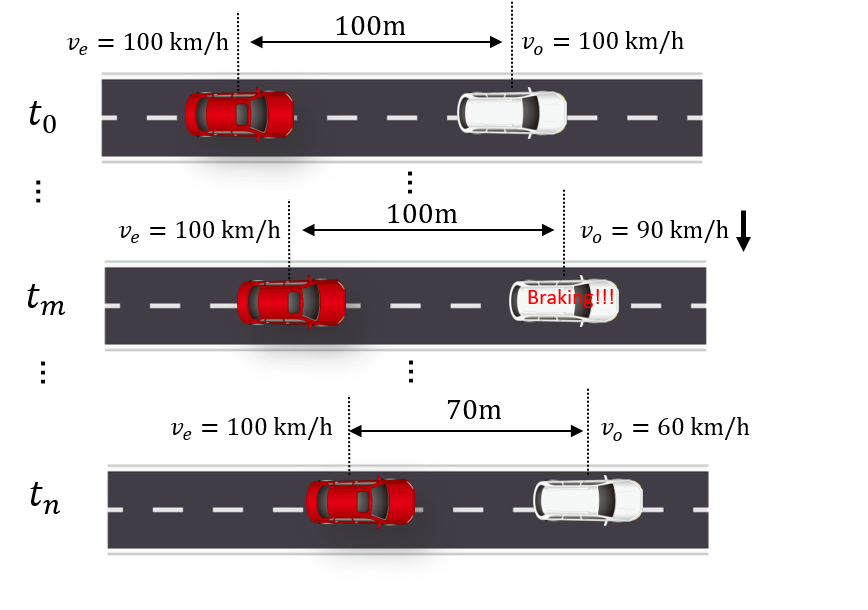}
    \caption{A schematic diagram illustrating the impact of motion information lag on practical applications.}
    \label{VDE_diag}
\end{figure}

To better illustrate the significance of the VDE metric, we provide a schematic diagram in Fig.\ref{VDE_diag}. The diagram shows two vehicles traveling at a speed of 100 kilometers per hour: the red one represents the autonomous vehicle, and the white one represents the obstacle ahead. The initial safe distance between the two vehicles is set at 100 meters. Suppose at time point \(t_m\), the leading vehicle begins to decelerate urgently and reduces its speed to 60 kilometers per hour by time point \(t_n\). If there is a delay in the autonomous vehicle's perception of the leading vehicle's speed, it might mistakenly believe that the leading vehicle is still traveling at 100 kilometers per hour. This can lead to an imperceptible reduction in the safe distance between the two vehicles. It is not until time point \(t_n\) that the autonomous vehicle finally perceives the deceleration of the leading vehicle, by which time the safe distance may be very close to the limit. Therefore, optimizing the motion information output by the multi-object tracking module is also crucial in autonomous driving.

% % AOE
% AOE represents the average error of the orientation angle, is calculated as:
% \begin{equation}
%   \text{AOE} = \text {Yaw}_{\mathrm{gt}} - \text {Yaw}_{\mathrm{dt}} + n \cdot 360^\circ \quad n \in \{-1, 0, 1\},
%   \label{eq:AOE}
% \end{equation}
% where $Yaw_{gt}$ and $Yaw_{dt}$ represent the true orientation angle and the orientation angle obtained from the tracking system, respectively($Yaw \in [0^\circ, 360^\circ]$).  Similar to the calculation method of Equation \eqref{eq:VAE}.

% % ALE、AWE 
% ALE and AWE respectively represent the average length error of the object's width and the average width error of the object box.
% \begin{equation}
%   \mathrm{ALE}=\left|l_{\mathrm{gt}}-l_{\mathrm{dt}}\right|, \quad \mathrm{AWE}=\left|w_{\mathrm{gt}}-w_{\mathrm{dt}}\right|,
%   \label{eq:ALE}
% \end{equation}
% where $l_{\mathrm{dt}}$ and $w_{\mathrm{dt}}$ represent the lengths and widths of the target boxes obtained by the tracking system, while $l_{\mathrm{gt}}$ and $w_{\mathrm{dt}}$ represent the true values of the target box lengths and widths.

\section{Experiment}
\label{sec:EXPERIMENT}

In this section, we first outline our experimental setup, including the datasets and implementation details. We then conduct a comprehensive comparison between our method and SOTA approaches on the 3D MOT benchmarks of the KITTI, nuScenes, and Waymo datasets. Following this, we evaluate our newly proposed dynamic metrics using various methods. Finally, we provide a series of ablation studies and related analyses to investigate the various design choices in our approach.

% 5.1
\subsection{Dataset and Implementation Details}

\paragraph{A. Datasets \vspace{5pt} \\}

 \textbf{KITTI:} The KITTI tracking benchmark \cite{geiger2012we} consists of 21 training sequences and 29 testing sequences. The training dataset comprises a total of 8,008 frames, with an average of 3.8 detections per frame, while the testing dataset contains 11,095 frames, with an average of 3.5 detections per frame. The point cloud data in KITTI is captured using a Velodyne HDL-64E LiDAR sensor, with a scan frequency of 10 Hz. The time interval $\delta$ between scans, used to infer actual velocity and acceleration, is 0.1 seconds. We compare our results on the vehicle category in the test dataset with those of other methods.

\textbf{NuScenes:} The nuScenes dataset \cite{caesar2020nuscenes} is a large-scale dataset that contains 1,000 driving sequences, each spanning 20 seconds. LiDAR data in nuScenes is provided at 20 Hz, but the 3D labels are only available at 2 Hz. The nuScenes dataset includes seven categories of data, and we evaluate all of these categories.

\textbf{Waymo:} The Waymo Open Dataset \cite{sun2020scalability} comprises 1,150 sequences, with 798 training sequences, 202 validation sequences, and 150 test sequences. Each sequence contains 20 seconds of continuous driving data within a range of [75m, 75m]. The 3D labels are provided for three categories: vehicles, pedestrians, and cyclists. We evaluate all categories in this dataset as well.

\paragraph{B. Implementation Details \vspace{5pt} \\}

Our method is fully implemented in Python on CPU, without the use of GPU acceleration. To ensure optimal performance, during the data preprocessing stage, we filter out bounding boxes with low detection scores and apply Non-Maximum Suppression (NMS) to remove those with significant overlap. We employ three Kalman filters to model the pose, size, and heading angle of the targets, respectively. For the cost calculation, we utilize our newly proposed Ro\_GDIoU. In our ablation studies, we compare the results obtained using different cost calculation methods. In the matching process, the first match uses the Hungarian algorithm, while the second match employs a greedy algorithm. The ablation study confirms the effectiveness of the double matching approach. In trajectory management, we set different lifecycles for different categories. For more detailed information on hyperparameter settings, please refer to our code implementation.

% \vspace{10pt}

% kitti 测试集
\begin{table*}[ht]
\captionsetup{justification=centering}
\caption{The comparison of the existing methods on the KITTI test set. The best performance is marked in \textbf{\textcolor{red}{red}}, and the second-best is marked in \textbf{\textcolor{blue}{blue}}.}
\centering
\label{KITTIResult}
\renewcommand{\arraystretch}{1.15} % 增加行间距
\resizebox{\textwidth}{!}{%
\begin{tabular}{lccccccccccccc}
\hline
\textbf{Method} & \textbf{Detector} & \textbf{Mode} & \textbf{HOTA\%$\uparrow$} & \textbf{AssA\%$\uparrow$} & \textbf{MOTA\%$\uparrow$} & \textbf{MOTP\%$\uparrow$} & \textbf{TP$\uparrow$} & \textbf{FP$\downarrow$} & \textbf{FN$\downarrow$} & \textbf{IDS$\downarrow$} & \textbf{FRAG$\downarrow$} \\
\hline
 \rowcolor{gray!20} TripletTrack \cite{marinello2022triplettrack}$(\textrm{CVPR}^{\prime} 22)$ & QD-3DT \cite{hu2022monocular} & online & 73.58 & 74.66 & 84.32 & 86.06 & 29750 & 4642 & \textbf{\textcolor{blue}{430}} & 322 & 522 \\

\rowcolor{gray!20} RAM \cite{tokmakov2022object} $(\textrm{ICML}^{\prime} 22)$ & CenterNet \cite{duan2019centernet} & online & 79.53 & 80.94 & \textbf{\textcolor{red}{91.61}} & 85.79 & \textbf{\textcolor{blue}{32298}} & \textbf{\textcolor{blue}{2094}} & 583 & 210 & \textbf{\textcolor{blue}{158}} \\

\rowcolor{gray!20} FNC2 \cite{jiang2023novel} $(\textrm{TIV}^{\prime} 23)$ & Voxel R-CNN \cite{deng2021voxel} & online & 73.19 & 73.77 & 84.21 & 85.86 & 31629 & 2763 & 2472 & 195 & 301 \\

\rowcolor{gray!20} OC-SORT \cite{cao2023observation} $(\textrm{CVPR}^{\prime} 23)$ & CenterNet \cite{duan2019centernet} & online & 76.54 & 76.39 & 90.28 & 85.53 & 31707 & 2685 & \textbf{\textcolor{red}{407}} & 250 & 280 \\

\rowcolor{gray!20} CAMO-MOT \cite{wang2023camo} $(\textrm{TITS}^{\prime} 23)$ & PointGNN \cite{shi2020point} & online & 79.95 & - & 90.38 & 85.00 & - & 2322 & 962 & \textbf{\textcolor{red}{23}} & - \\

\rowcolor{gray!20} LEGO \cite{zhang2023lego} $(\textrm{arxiv}^{\prime} 23)$ & VirConv \cite{wu2023virtual} & online & \textbf{\textcolor{blue}{80.75}} & \textbf{\textcolor{blue}{83.27 }}& \textbf{\textcolor{blue}{90.61}} & \textbf{\textcolor{blue}{86.66}} & \textbf{\textcolor{red}{32823}} & \textbf{\textcolor{red}{1569}} & 1445 & 214 & \textbf{\textcolor{red}{109}} \\

\rowcolor{gray!20} PNAS-MOT \cite{peng2024pnas} $(\textrm{RAL}^{\prime} 24)$ & - & online & 67.32 & 58.99 & 89.59 & 85.44 & 32131 & 2261 & 568 & 751 & 276 \\

\rowcolor{gray!20} SpbTracker \cite{im2024spb3dtracker} $(\textrm{arxiv}^{\prime} 24)$ & DSVT \cite{wang2023dsvt} & online & 72.66 & 71.43 & 86.51 & 86.07 & 30884 & 3508 & 875 & 257 & 496 \\

\rowcolor{gray!20} UG3DMOT \cite{he20243d} $(\textrm{SP}^{\prime} 24)$ & CasA \cite{wu2022casa} & online & 78.60 & 82.28 & 87.98 & 86.56 & 31399 & 2993 & 1111 & \textbf{\textcolor{blue}{30}} & 360 \\

\hline
\rowcolor{gray!20} Ours & VirConv \cite{wu2023virtual} & online & \textbf{\textcolor{red}{80.78}} & \textbf{\textcolor{red}{84.30}} & 89.82 & \textbf{\textcolor{red}{86.71}} & 32207 & 2185 & 1252 & 64 & 438 \\
\hline
CasTrack \cite{wu2022casa} $(\textrm{CVPR}^{\prime} 22)$ & CasA \cite{wu2022casa} & offline & 81.00 & 84.22 & \textbf{\textcolor{red}{91.91}} & 86.08 & \textbf{\textcolor{blue}{32859}} & \textbf{\textcolor{blue}{1533}} & 1227 & 24 & 107 \\
Rethink MOT \cite{wang2023towards} $(\textrm{ICRA}^{\prime} 23)$ & PointRCNN \cite{shi2019pointrcnn} & offline & 80.39 & 83.64 & \textbf{\textcolor{blue}{91.53}} & 85.58 & \textbf{\textcolor{red}{33094}} & \textbf{\textcolor{red}{1298}} & 1569 & 46 & 134 \\
VirConvTrack \cite{wu2023virtual} $(\textrm{CVPR}^{\prime} 23)$ & VirConv \cite{wu2023virtual} & offline & 81.87 & \textbf{\textcolor{blue}{86.39}} & 90.24 & \textbf{\textcolor{blue}{86.82}} & 31744 & 2648 & \textbf{\textcolor{blue}{702}} & \textbf{\textcolor{red}{8}} & \textbf{\textcolor{blue}{77}} \\
BiTrack \cite{huang2024bitrack} $(\textrm{arxiv}^{\prime} 24)$ & VirConv \cite{wu2023virtual} & offline & \textbf{\textcolor{blue}{82.39}} & 85.57 & 91.52 & \textbf{\textcolor{red}{87.55}} & 32445 & 1947 & 948 & 20 & 270 \\
\hline
% Ours & VirConv & online & 80.78 & 84.29 & 89.82 & 86.71 & 32204 & 2188 & 1248 & 65 & 439 \\
Ours & VirConv \cite{wu2023virtual} & offline & \textbf{\textcolor{red}{82.56}} & \textbf{\textcolor{red}{86.64}} & 91.62 & \textbf{\textcolor{blue}{86.82}} & 32064 & 2328 & \textbf{\textcolor{red}{542}} & \textbf{\textcolor{blue}{12}} & \textbf{\textcolor{red}{59}} \\
\hline
\end{tabular}%
}
\end{table*}

% nuScenes 测试集
\begin{table*}[ht]
\captionsetup{justification=centering}
\caption{Comparison of existing methods on the nuScenes test set. The best performance is marked in \textbf{\textcolor{red}{red}}, and the second-best in \textbf{\textcolor{blue}{blue}}. Here, (Bic., Motor, Ped., Tra., Tru.) denote (Bicycle, Motorcycle, Pedestrian, Trailer, Truck), and (CR, FC) refer to (Cascade R-CNN \cite{cai2018cascade}, FocalsConv \cite{chen2022focal}).}
\centering
\label{nuScenesResult}
\renewcommand{\arraystretch}{1.15} % 增加行间距
\renewcommand{\arraystretch}{1.15} % 增加行间距
\resizebox{\textwidth}{!}{%
\begin{tabular}{lccccccccccccc}
\hline
\multirow{2}{*}{\textbf{Method}} & \multirow{2}{*}{\textbf{Detector}} & \multicolumn{8}{c}{\textbf{AMOTA}\%$\uparrow$} & \multirow{2}{*}{\textbf{TP$\uparrow$}} & \multirow{2}{*}{\textbf{FP$\downarrow$}} & \multirow{2}{*}{\textbf{FN$\downarrow$}} & \multirow{2}{*}{\textbf{IDS}$\downarrow$} \\
\cline{3-10}
 &  & \textbf{Overall} & \textbf{Bic.} & \textbf{Bus} & \textbf{Car} & \textbf{Motor} & \textbf{Ped.} & \textbf{Tra.} & \textbf{Tru.} &  &  &  &  \\
\hline
DeepFusionMOT \cite{wang2022deepfusionmot} $(\textrm{RAL}^{\prime} 22)$ & CenterPoint \cite{yin2021center} \& CR \cite{cai2018cascade} & 63.5 & 52.0 & 70.8 & 72.5 & 69.6 & 55.4 & 64.9 & 59.6 & 84304 & 19303 & 33556 & 1075 \\

SimpleTrack \cite{pang2022simpletrack} $(\textrm{ECCV}^{\prime} 22)$ & CenterPoint \cite{yin2021center} & 66.8 & 40.7 & 71.5 & 82.3 & 67.4 & 79.6 & 67.3 & 58.7 & 95539 & 17514 & 24351 & 575 \\

MoMA-M3T \cite{huang2023delving} $(\textrm{CVPR}^{\prime} 23)$ & PGD \cite{wang2022probabilistic} & 42.5 & 34.1 & 39.5 & 63.4 & 46.4 & 46.9 & 33.0 & 34.6 &  73495 & 18563 & 42934 & 3136 \\

3DMOTFormer \cite{ding20233dmotformer} $(\textrm{ICCV}^{\prime} 23)$ & CenterPoint \cite{yin2021center} & 68.2 & 37.4 & 74.9 & 82.1 & 70.5 & 80.7 & 69.6 & 62.6 &  95790 & 18322 & 23337 & 438 \\

VoxelNeXt  \cite{chen2023voxelnext} $(\textrm{CVPR}^{\prime} 23)$ & VoxelNeXt \cite{chen2023voxelnext} & 71.0 & 52.6 & 74.7 & 82.6 & 73.1 & 76.0 & 73.8 & 64.4 & 97075 & 18348 & 21836 & 654 \\

FocalFormal3D-F \cite{chen2023focalformer3d} $(\textrm{ICCV}^{\prime} 23)$ & FocalFormer3D \cite{chen2023focalformer3d} & 73.9 & 54.1 & \textbf{\textcolor{red}{79.2}} & 83.4 & 74.6 & 84.1 & 75.2 & 66.9 & 97987 & \textbf{\textcolor{blue}{15547}} & 20754 & 824 \\

MSMDFusion \cite{jiao2023msmdfusion} $(\textrm{CVPR}^{\prime} 23)$ & MSMDFusion \cite{jiao2023msmdfusion} & 74.0 & 57.4 & 76.7 & 84.9 & 75.4 & 80.7 & 75.4 & 67.1 & 98624 & \textbf{\textcolor{red}{14789}} & 19853 & 1088 \\

BEVFusion \cite{liu2023bevfusion} $(\textrm{ICRA}^{\prime} 23)$ & BEVFusion \cite{liu2023bevfusion} & 74.1 & 56.0 & 77.9 & 83.1 & 74.8 & 83.7 & 73.4 & \textbf{\textcolor{red}{69.5}} & 99664 & 19997 & 19395 & 506 \\

CAMO-MOT \cite{wang2023camo} $(\textrm{TITS}^{\prime} 23)$ & BEVFusion \cite{liu2023bevfusion} \& FC \cite{chen2022focal} & 75.3 & \textbf{\textcolor{red}{59.2}} & 77.7 & 85.8 & 78.2 & \textbf{\textcolor{red}{85.8}} & 72.3 & \textbf{\textcolor{blue}{67.7}} & 101049 & 17269 & 18192 & 324 \\

Poly-MOT \cite{li2023poly} $(\textrm{ICRA}^{\prime} 23)$ & LargeKernel3D \cite{chen2023largekernel3d} & 75.4 & 58.2 & \textbf{\textcolor{blue}{78.6}} & \textbf{\textcolor{blue}{86.5}} & 81.0 & 82.0 & 75.1 & 66.2 & \textbf{\textcolor{blue}{101317}} & 19673 & \textbf{\textcolor{blue}{17956}} & \textbf{\textcolor{blue}{292}} \\

CR3DT \cite{baumann2024cr3dt} $(\textrm{arxiv}^{\prime} 24)$ & CR3DT \cite{baumann2024cr3dt} & 35.5 & 23.7 & 30.3 & 56.9 & 42.1 & 33.9 & 27.2 & 34.2 & 62921 & 14428 & 54716 & 1928 \\

ADA-Track \cite{ding2024ada} $(\textrm{arxiv}^{\prime} 24)$ & DETR3D \cite{wang2022detr3d} & 45.6 & 33.4 & 38.2 & 66.4 & 48.4 & 53.4 & 43.7 & 35.9 & 79051 & 15699 & 39680 & 834 \\

ShaSTA \cite{sadjadpour2023shasta} $(\textrm{RAL}^{\prime} 24)$ & CenterPoint \cite{yin2021center} & 69.6 & 41.0 & 73.3 & 83.8 & 72.7 & 81.0 & 70.4 & 65.0 & 97799 & 16746 & 21293 & 473 \\

NEBP-V3 \cite{liang2023neural} $(\textrm{TSP}^{\prime} 24)$ & CenterPoint \cite{yin2021center} \& FC \cite{chen2022focal} & 74.6 & 49.9 & \textbf{\textcolor{blue}{78.6}} & 86.1 & 80.7 & 83.4 & 76.1 & 67.3 & 99872 & 17243 & 19316 & 377 \\

Fast-Poly \cite{li2024fast} $(\textrm{arxiv}^{\prime} 24)$ & LargeKernel3D \cite{chen2023largekernel3d} & \textbf{\textcolor{blue}{75.8}} & 57.3 & 76.7 & 86.3 & \textbf{\textcolor{red}{82.6}} & \textbf{\textcolor{blue}{85.2}} & \textbf{\textcolor{blue}{76.8}} & 65.6 & 100824 & 17098 & 18415 & 326 \\

\hline
Ours & LargeKernel3D \cite{chen2023largekernel3d} & \textbf{\textcolor{red}{76.3}} & \textbf{\textcolor{blue}{59.1}} & 77.2 & \textbf{\textcolor{red}{87.9}} & \textbf{\textcolor{blue}{81.9}} & 83.2 & \textbf{\textcolor{red}{77.9}} & 66.7 & \textbf{\textcolor{red}{103327}} & 19643 & \textbf{\textcolor{red}{15996}} & \textbf{\textcolor{red}{242}} \\
\hline
\end{tabular}%
}
\end{table*}

% Waymo 测试集
\begin{table*}[!t]
\captionsetup{justification=centering}
\caption{The comparison of the existing algorithms on the Waymo test set. The best performance is marked in \textbf{\textcolor{red}{red}}, and the second-best is marked in \textbf{\textcolor{blue}{blue}}.}
\centering
\label{WaymoResult}
\resizebox{\textwidth}{!}{%
\begin{tabular}{lcccccccccc}
\hline
\multirow{2}{*}{\textbf{Method}} & \multirow{2}{*}{\textbf{Detector}} & \multicolumn{4}{c}{\textbf{MOTA}\%$\uparrow$} & \multirow{2}{*}{\textbf{MOTP\%$\downarrow$}} & \multirow{2}{*}{\textbf{FP\%$\downarrow$}} & \multirow{2}{*}{\textbf{Mismatch\%$\downarrow$}} & \multirow{2}{*}{\textbf{Miss\%$\downarrow$}} \\
\cline{3-6}
 &  & \textbf{Overall} & \textbf{Vehicle} & \textbf{Pedestrian} & \textbf{Cyclist} &  &  &  &  \\
\hline
CenterPoint \cite{yin2021center} $(\textrm{CVPR}^{\prime} 21)$ & CenterPoint \cite{yin2021center} & 58.67 & 59.38 & 56.64 & 60.00 & 25.05 & 9.93 & 0.72 & 30.68 \\
ImmortalTracker \cite{wang2021immortal} $(\textrm{arxiv}^{\prime} 21)$ & CenterPoint \cite{yin2021center} & 60.92 & 60.55 & 60.60 & 61.61 & 24.94 & 9.57 & 0.10 & 29.41 \\
SimpleTrack \cite{pang2022simpletrack} $(\textrm{ECCV}^{\prime} 22)$ & CenterPoint \cite{yin2021center} & 60.18 & 60.30 & 60.13 & 60.12 & 24.91 & 9.68 & 0.38 & 29.57 \\
CasTrack \cite{wu2022casa} $(\textrm{TGRS}^{\prime} 22)$ & CasA \cite{wu2022casa} & 62.60 & 63.66 & 64.79 & 59.34 & 23.78 & 9.28 & 0.13 & 27.99 \\
YONTD\textunderscore MOT \cite{wang2023you} $(\textrm{arxiv}^{\prime} 23)$ & CenterPoint \cite{yin2021center} & 55.70 & 60.48 & 50.08 & 56.55 & 24.07 & \textbf{\textcolor{blue}{8.44}} & 0.60 & 35.26 \\
TrajectoryFormer \cite{chen2023trajectoryformer} $(\textrm{ICCV}^{\prime} 23)$ & MPPNet \cite{chen2022mppnet} & 66.09 & 65.73 & 66.15 & 66.38 & 23.60 & 10.06 & 0.38 & 23.47 \\
CTRL\textunderscore FSD\textunderscore TTA \cite{fan2023once} $(\textrm{ICCV}^{\prime} 23)$ & CTRL \cite{fan2023once} & 73.29 & 74.29 & 74.21 & \textbf{\textcolor{blue}{71.37}} & 22.80 & 8.45 & \textbf{\textcolor{red}{0.04}} & 18.21 \\
DetZero \cite{ma2023detzero} $(\textrm{ICCV}^{\prime} 23)$ & DetZero \cite{ma2023detzero} & \textbf{\textcolor{red}{75.05}} & \textbf{\textcolor{red}{75.97}} & \textbf{\textcolor{red}{76.03}} & \textbf{\textcolor{red}{73.16}} & \textbf{\textcolor{red}{22.24}} & 
\textbf{\textcolor{red}{7.70}} & 0.07 & \textbf{\textcolor{red}{17.17}} \\
Fast-Poly \cite{li2024fast} $(\textrm{arxiv}^{\prime} 24)$ & CasA \cite{wu2022casa} & 62.77 & 63.06 & 64.92 & 62.77 & 23.74 & 8.71 & 0.12 & 27.59 \\
\hline
Ours & CTRL \cite{fan2023once} & \textbf{\textcolor{blue}{73.44}} & \textbf{\textcolor{blue}{74.65}} & \textbf{\textcolor{blue}{74.37}} & 71.30 & \textbf{\textcolor{blue}{22.78}} & 8.47 & \textbf{\textcolor{blue}{0.05}} & \textbf{\textcolor{blue}{18.04}} \\
\hline
\end{tabular}%
}
\end{table*}

% 5.2
\subsection{Quantitative Experiment}
We compared MCTrack with published and peer-reviewed SOTA methods on the test sets of the KITTI, nuScenes, and Waymo datasets. Our method demonstrated superior performance across these datasets. Next, we will provide a detailed description of the experimental results on each dataset.

\textbf{KITTI:} On the KITTI dataset, MCTrack demonstrated outstanding performance in both online and offline testing, achieving HOTA scores of 80.78\% and 82.46\% respectively, as shown in TABLE \ref{KITTIResult}. These scores are leading among all tested methods. Notably, MCTrack excelled in Association Accuracy (AssA) with a score of 86.55\% and also displayed the lowest rate of False Negatives (FN). The AssA metric is designed to evaluate the precision of association tasks. Securing the top position in the rankings with our AssA score is a testament to MCTrack's exceptional ability to accurately match and connect detection targets with high fidelity.

Furthermore, online tracking performance is particularly crucial in practical engineering applications, as it involves real-time processing and usually does not include subsequent trajectory optimization. In this respect, MCTrack also performed exceptionally well, with its online tracking capabilities being the best among all methods compared.

\textbf{NuScenes:} On the nuScenes dataset, MCTrack achieved an AMOTA score of 76.3\%, the best performance among all participating 3D multi-object tracking systems. As shown in TABLE \ref{nuScenesResult}. Notably, MCTrack demonstrated superior tracking results in key detection categories such as car and trailer, outperforming other tracking systems. Additionally, for the Kalman filter, we employed only a simple Constant Velocity model. Moreover, MCTrack achieved the highest number of TP and the lowest number of FN and IDS. This result demonstrates MCTrack's exceptional performance in maintaining tracking stability.

\textbf{Waymo:} In the Waymo dataset, our method outperforms others when using a unified detector as as show in TABLE \ref{WaymoResult}. Although MCTrack rank second in the leaderboard, it's important to note that the detector used by DetZero \cite{ma2023detzero}, the top-ranked method, significantly outperforms ours in various metrics, such as a higher mean Average Precision (mAP) by more than two points. We believe that the methods, not only ours but also those of all other ranked methods, are not directly comparable.

It is particularly noteworthy that the tracking results we obtained across all three datasets were achieved using the same baseline framework. This fully demonstrates that our baseline framework and methodology not only possess high robustness but also display clear superiority.

% 5.3
\subsection{Motion Metrics Evaluation}

In Section \ref{sec:METRICS}, we discussed the limitations of existing 3D multi-object tracking metrics and introduced a series of motion metrics. Here, we analyze the results of these motion metrics across different methods. The experiments were conducted on the nuScenes dataset, which includes ground truth speed in its annotations. We used these ground truth values as the reference standard for evaluating our proposed motion metrics. The methods compared include detected speed, speed from differentiation, curve-fitted speed, and Kalman filter-based speed estimation. The comparison results are shown in TABLE \ref{evalvelocity}.

The velocity obtained through differentiation is calculated based on the change in position and the time difference. The calculation formula is as follows:

\begin{equation}
    v_{\textrm{diff}} = \frac{(x_{\textrm{cur}} - x_{\textrm{prev}}, y_{\textrm{cur}} - y_{\textrm{prev}})}{\Delta t},
    \label{eq:diff}
\end{equation}
where $(x_\textrm{prev}, y_\textrm{prev })$ represents the position of the previous frame, $(x_\textrm{cur}, y_\textrm{cur})$ represents the position of the current frame, and $\Delta t$ represents the time difference between the two frames.

Curve fitting is based on the positions of the most recent three frames, assuming that the position changes linearly with time. The velocity in each direction is calculated through linear fitting. The linear fitting function is defined as:

\begin{equation}
    p(x)=a \cdot x+b,
  \label{eq:diff}
\end{equation}
$a$ is the slope of the fitted line, which represents the velocity. $b$ is the intercept.

For the $x$ and $y$ coordinates of the position, we perform fitting for frame number $\left[f_{n-2}, f_{n-1}, f_{n}\right]$ and position coordinates $\left[x_{n-2}, x_{n-1}, x_{n}\right]$ and $\left[y_{n-2}, y_{n-1}, y_{n}\right]$, and the resulting velocity is:

\begin{equation}
    \mathbf{v}_{\text {curve }}=\left(a_{x}, a_{y}\right).
  \label{eq:curve}
\end{equation}

The results show that the Kalman filter achieved the lowest VAE and VNE, indicating that it has the highest accuracy in terms of both speed magnitude and direction. Additionally, the Kalman filter also recorded the lowest VIR, demonstrating its ability to effectively suppress speed reversal fluctuations. As expected, the differentiation method resulted in the lowest VDE, indicating the fastest speed response. The curve fitting method, on the other hand, produced the smallest VSE, meaning it generated a smoother speed curve, but at the cost of a larger VSE. As for VAIE, it is difficult to judge its quality solely based on its value, and it typically requires evaluation in the context of practical engineering applications.

% 动态指标度量
% \begin{table*}[!ht]
% \captionsetup{justification=centering}
% \caption{The comparison of dynamic metric results obtained by different methods.}
% \centering
% \label{evalvelocity}
% \renewcommand{\arraystretch}{1.3} % 增加行间距
% \resizebox{\textwidth}{!}{%
% \begin{tabular}{lccccccccccccc}
% \hline
% \textbf{Method} & \textbf{TP$\uparrow$} & \textbf{VAE(\textdegree)$\downarrow$} & \textbf{VNE(m/s)$\downarrow$} & \textbf{VDE(s)$\downarrow$} & \textbf{VSE(m/s)$\downarrow$} & \textbf{VAIE(\textdegree)} & \textbf{VIR(\%)$\downarrow$} & \textbf{AOE(\textdegree)$\downarrow$} & \textbf{AWE(m)$\downarrow$} & \textbf{ALE(m)$\downarrow$} \\
% \hline
% Detection & 86721 & 6.50 & 0.59 & 3.78 & 1.49 & 138.71 & 0.07 & 1.01 & \textbf{0.15} & 2.31 \\
% Differentiation & 86721 & 8.13 & 0.83 & \textbf{3.73} & 1.63 & 136.92 & 0.08 & - & - & - \\
% Curve Fitting & 86721 & 8.64 & 2.48 & 4.33 & \textbf{0.73} & 136.83 & 0.10 & - & - & - \\
% Kalman Filter & 86718 & \textbf{6.25} & \textbf{0.55} & 3.76 & 1.47 & 140.32 & \textbf{0.06} & 1.43 & 0.15 & 2.31 \\
% \hline
% \end{tabular}%
% }
% \end{table*}
\begin{table}[!h]
\captionsetup{justification=centering}
\caption{The comparison of motion metric results obtained by different methods.}
\centering
\label{evalvelocity}
\renewcommand{\arraystretch}{1.3} % 增加行间距
\resizebox{\columnwidth}{!}{% 调整到半边栏宽度
\begin{tabular}{lccccccc}
\hline
\textbf{Method} & \textbf{TP$\uparrow$} & \textbf{VAE(\textdegree)$\downarrow$} & \textbf{VNE(m/s)$\downarrow$} & \textbf{VDE(s)$\downarrow$} & \textbf{VSE(m/s)$\downarrow$} & \textbf{VAIE(\textdegree)} & \textbf{VIR(\%)$\downarrow$} \\
\hline
Detection   & 88791      & 6.50      & 0.60      & 3.82      & 1.47       & 138.94     & 0.07 \\
Differentiation & 88791     & 8.03      & 0.85       &  \textbf{3.76}     & 1.65     & 137.44      & 0.08 \\
Curve Fitting   & 88791                & 8.48                           & 2.48                      & 4.35                       &  \textbf{0.74}                       & 137.55                     & 0.10 \\
Kalman Filter   & 88692                &  \textbf{6.30}                           &  \textbf{0.56}                      & 3.78                       & 1.46                       & 141.04                     &  \textbf{0.06} \\
\hline
\end{tabular}%
}
\end{table}

\begin{figure}[h]
    \centering
    \includegraphics[width=1\linewidth]{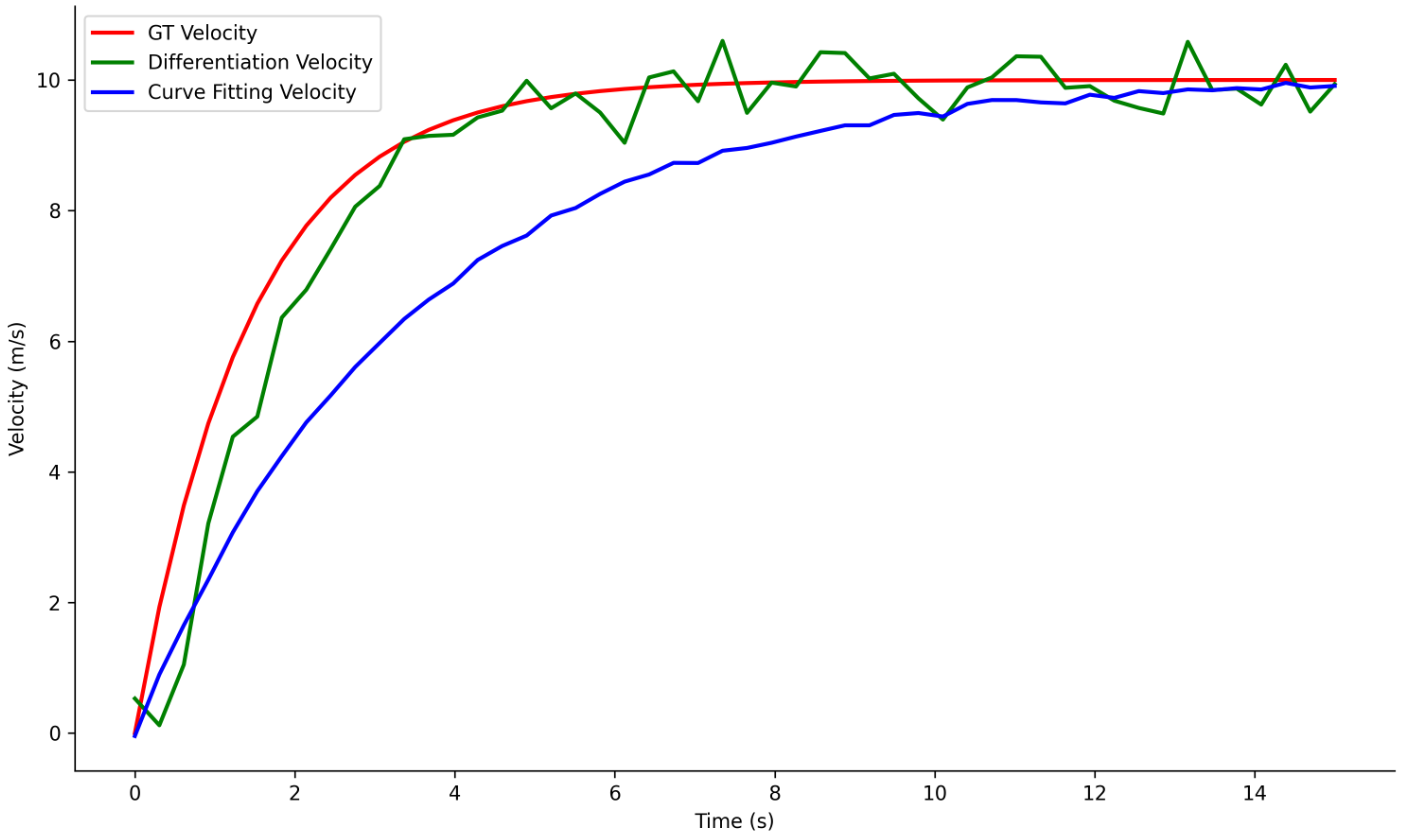}
    \caption{Comparison of velocity curves from different methods.}
    \vspace{-10pt}
    \label{eval_velocity}
\end{figure}

To better explain the meaning of the VDE and VSE metrics, we have created a diagram, as shown in Fig.\ref{eval_velocity}. The green curve represents the ground truth velocity, the red curve shows the velocity obtained through differencing, and the blue curve represents the velocity obtained through curve fitting. Compared to curve fitting, the differenced velocity has a faster response, but the smoothness of the curve is inferior. The experimental results also indicate that while the differencing method provides a quicker response, curve fitting generates a much smoother curve.

% 5.4
\subsection{Ablation Studies}
To validate the effectiveness of the components in MCTrack, we conducted comprehensive ablation experiments on three datasets: KITTI, nuScenes, and Waymo. For the KITTI dataset, the ablation experiments were performed only on the car category, while for the nuScenes and Waymo datasets, the experiments covered all categories. Our ablation study is divided into two main parts: the first part involves conducting ablation experiments on Ro\_GDIoU and secondary matching within our unified framework; the second part integrates our Ro\_GDIoU matching method into other state-of-the-art (SOTA) methods for comparison experiments.

\subsubsection{Pipeline Ablation Studies}
\paragraph{A. Ro\_GDIoU  \vspace{5pt} \\}  
We conducted a series of ablation experiments based on the Poly-MOT \cite{li2023poly} and PC3T \cite{han2023pc3t} methods on the nuScenes and KITTI datasets, where we replaced the Ro\_GDIoU in MCTrack with GIoU and DIoU, respectively, to demonstrate the effectiveness and superiority of our cost method. This comparative experiment effectively proves the performance advantages of the proposed method. 

% 表5和6超出范围了，调整

\begin{table}[!h]
\captionsetup{justification=centering}
\caption{Comparison of results using different cost calculations on MCTrack with the nuScenes dataset (using the CenterPoint detector \cite{yin2021center}).}
\centering
\label{nuScenescompare}
\renewcommand{\arraystretch}{1.3} % 增加行间距
\resizebox{0.5\textwidth}{!}{%
\begin{tabular}{lccccc}
% {p{3.5cm}>{\centering\arraybackslash}p{1.0cm}>{\centering\arraybackslash}p{1.0cm}>{\centering\arraybackslash}p{1.0cm}>{\centering\arraybackslash}p{1.0cm}>{\centering\arraybackslash}p{1.0cm}}
\hline
\textbf{Method} & \textbf{AMOTA$\uparrow$} & \textbf{MOTA$\uparrow$}  &  \textbf{TP$\uparrow$} & \textbf{FP$\downarrow$} & \textbf{IDS$\downarrow$} \\
\hline
MCTrack + GIoU   & 73.7 & 63.5 & 84430 & 13851 & \textbf{233}   \\
MCTrack + DIoU   & 73.1 & 62.9 & 85231 & 13332 & 240 \\
MCTrack + Ro\_GDIoU  & \textbf{74.0} & \textbf{64.0} & \textbf{85900} & \textbf{13083} & 275  \\
\hline
\end{tabular}%
}
\end{table}

\begin{table}[!h]
\captionsetup{justification=centering}
\caption{Comparison of the best results using different cost calculations on MCTrack with the KITTI training dataset (using the VirConv detector \cite{wu2022casa}).}
\centering
\label{KITTIcompare}
\renewcommand{\arraystretch}{1.3} % 增加行间距
\resizebox{0.5\textwidth}{!}{%
\begin{tabular}{lccccc}
% {p{3.5cm}>{\centering\arraybackslash}p{1.0cm}>{\centering\arraybackslash}p{1.0cm}>{\centering\arraybackslash}p{1.0cm}>{\centering\arraybackslash}p{1.0cm}>{\centering\arraybackslash}p{1.0cm}}
\hline
{\textbf{Method}} & {\textbf{HOTA}$\uparrow$} & {\textbf{MOTA}$\uparrow$}& {\textbf{FP$\downarrow$}} & {\textbf{FN$\downarrow$}} & {\textbf{IDS}$\downarrow$} \\
\hline
MCTrack + GIoU & 83.55 & 86.17 & 1281 & 2042 & 5 \\
MCTrack + DIoU & 83.69 & 86.37 & 1297  & 1989 & \textbf{3} \\
MCTrack + Ro\_GDIoU & \textbf{83.90} & \textbf{86.42} & \textbf{1261} & \textbf{1958} & \textbf{3} \\
\hline
\end{tabular}%
}
\end{table}
% 表56字体格式检查

In the experiments on the nuScenes dataset, using Ro\_GDIoU resulted in improvements of 0.3\% and 0.9\% compared to GIoU and DIoU, respectively, fully demonstrating the effectiveness of the Ro\_GDIoU cost calculation strategy. However, in the experiments on the KITTI dataset, although using Ro\_GDIoU also improved the HOTA metric compared to GIoU and DIoU, the improvement was not as significant as in the nuScenes dataset. We speculate that this is due to the higher detection accuracy and relatively simpler scenarios in the KITTI dataset, leading to smaller improvements in tracking performance when using Ro\_GDIoU.

\paragraph{B. Secondary Matching \vspace{5pt} \\}
In practical engineering, obstacles directly in front of a vehicle typically have a much greater impact on driving safety compared to those in other directions. Therefore, to improve efficiency, we project only the obstacles ahead onto the RV plane for secondary matching. Table \ref{KITTIcompare_rv} demonstrates the effectiveness of RV matching on the KITTI dataset. Since these detectors are all based on LiDAR, their depth detection is relatively accurate, and thus RV matching does not lead to significant performance improvements. However, it does improve metrics such as FP and identity IDSW. Interestingly, we observed that the poorer the detector's performance, the more pronounced the enhancement brought by RV matching, while for the best detectors on the KITTI dataset, the benefits are quite limited. In real-world engineering applications, due to the limited computational resources of autonomous vehicles, their perception performance often falls short of what is demonstrated in open-source datasets. Therefore, we believe that RV matching technology can enhance perception performance in practical scenarios.

\begin{table}[!h]
\captionsetup{justification=centering}
\caption{Ablation experiments of secondary matching based on RV across different detectors, where BEV refers to matching on the BEV plane and RV refers to matching on the RV plane.}
\centering
\label{KITTIcompare_rv}
\renewcommand{\arraystretch}{1.3} % 增加行间距
\resizebox{\columnwidth}{!}{%
\begin{tabular}{lcccccc}
\toprule % 加粗顶部横线
\textbf{Detector} & \textbf{BEV} & \textbf{RV} & \textbf{MOTA}$\uparrow$ & \textbf{FN}$\downarrow$ & \textbf{FP}$\downarrow$ & \textbf{IDS}$\downarrow$ \\
\bottomrule % 加粗底部横线
PR \cite{shi2019pointrcnn} & \checkmark & \texttimes & 63.5 & \textbf{2467} & 6240 & 65\\
PR \cite{shi2019pointrcnn} & \checkmark & \checkmark & \textbf{64.2} & 2488 & \textbf{6087} & \textbf{60}\\
\midrule
SECOND \cite{yan2018second} & \checkmark & \texttimes & 79.3 & \textbf{4216} & 715 & 50\\
SECOND \cite{yan2018second} & \checkmark & \checkmark & \textbf{79.5} & 4219 & \textbf{695} & \textbf{45}\\
\midrule
CasA \cite{wu2022casa} & \checkmark & \texttimes & 83.9 & \textbf{1933} & 1902 & 30\\
CasA \cite{wu2022casa} & \checkmark & \checkmark & \textbf{84.1} & 1937 & \textbf{1874} & \textbf{29}\\
\midrule
VirConv \cite{wu2023virtual} & \checkmark & \texttimes & 85.9 & 2097 & 1279 & 24\\
VirConv \cite{wu2023virtual} & \checkmark & \checkmark & 85.9 & 2097 & \textbf{1276} & 24\\
\bottomrule % 加粗表格底部横线
\end{tabular}%
}
\end{table}

\subsubsection{Ro\_GDIoU for other methods}

To further validate the effectiveness of the proposed Ro\_GDIoU, we integrated it into two SOTA tracking methods, Poly-MOT and PC3T, which are widely used in the nuScenes and KITTI open-source communities. By incorporating Ro\_GDIoU into these established frameworks, we aimed to assess its impact on improving tracking accuracy and robustness across different datasets and real-world scenarios. The integration allows for a more comprehensive evaluation of Ro\_GDIoU's performance, demonstrating its potential to enhance the precision of object tracking in challenging environments. The results of these experiments are presented in TABLE \ref{nuScenescompare_GDIoU} and TABLE \ref{KITTIcompare_GDIoU}.

% nuScenes数据集的gdiou实验
\begin{table}[!h]
\captionsetup{justification=centering}
\caption{Comparison of results on the nuScenes dataset after replacing Poly-MOT's cost calculation with Ro\_GDIoU.}
\centering
\label{nuScenescompare_GDIoU}
\renewcommand{\arraystretch}{1.3} % 增加行间距
\resizebox{\columnwidth}{!}{%
\begin{tabular}{lcccccc}
\toprule % 加粗顶部横线
\textbf{Method} &  w/ \textbf{Ro} & \textbf{AMOTA}\%$\uparrow$ & \textbf{MOTA}\%$\uparrow$ & \textbf{TP}$\uparrow$ & \textbf{FP}$\downarrow$ & \textbf{IDS}$\downarrow$ \\
\bottomrule % 加粗底部横线
PolyMOT \cite{li2023poly}  & \texttimes & 73.1 & 61.9 & 84072 & \textbf{13051} & \textbf{232} \\
PolyMOT \cite{li2023poly}  & \checkmark & \textbf{73.5} & \textbf{63.0} & \textbf{84624} & 13164 & 279 \\

\bottomrule % 加粗表格底部横线
\end{tabular}%
}
\end{table}

% KITTI数据集的gdiou实验
% \begin{table}[!h]
% \captionsetup{justification=centering}
% \caption{Comparison of results on the KITTI training dataset after replacing PC3T's cost calculation with Ro\_GDIoU}
% \centering
% \label{KITTIcompare_GDIoU}
% \renewcommand{\arraystretch}{1.3} % 增加行间距
% \resizebox{\columnwidth}{!}{%
% \begin{tabular}{lcccccc}
% \toprule % 加粗顶部横线
% \textbf{Method}  & w/ \textbf{Ro} & {\textbf{HOTA}\%$\uparrow$} & {\textbf{MOTA}\%$\uparrow$} & {\textbf{FP$\downarrow$}} & {\textbf{FN$\downarrow$}} & {\textbf{IDS}$\downarrow$} \\
% \bottomrule % 加粗底部横线
% PC3T \cite{wu20213d}  & \texttimes & 83.17 & 86.07 & 1348 & \textbf{1992} & 13 \\
% PC3T \cite{wu20213d}  & \checkmark & \textbf{83.65} & \textbf{86.32} & \textbf{1252} & 2038 & \textbf{3} \\

% \midrule
% DFMOT \cite{wang2022deepfusionmot}  & \texttimes & 77.45 & 87.27 & \textbf{643} & 2337 & 76 \\
% DFMOT \cite{wang2022deepfusionmot} & \checkmark & \textbf{77.76} & \textbf{87.29} & 724  & \textbf{2282} & \textbf{54}  \\

% \midrule
% EagerMOT \cite{kim2021eagermot} & \texttimes & 76.51 & 85.02 & 1239 & 2219 & 147 \\
% EagerMOT \cite{kim2021eagermot} & \checkmark & \textbf{76.57} & \textbf{85.07} & 1239 & 2219 & \textbf{137} \\

% \midrule
% OC-SORT \cite{cao2023observation}  & \texttimes & -- & -- & -- & -- & -- \\
% OC-SORT \cite{cao2023observation} & \checkmark & -- & -- & -- & -- & -- \\

% \bottomrule % 加粗表格底部横线
% \end{tabular}%
% }
% \end{table}

\begin{table}[!h]
\captionsetup{justification=centering}
\caption{Comparison of results on the KITTI training dataset after replacing PC3T's cost calculation with Ro\_GDIoU.}
\centering
\label{KITTIcompare_GDIoU}
\renewcommand{\arraystretch}{1.3} % 增加行间距
\resizebox{\columnwidth}{!}{%
\begin{tabular}{lcccccc}
\toprule % 加粗顶部横线
\textbf{Method}  & w/ \textbf{Ro} & {\textbf{HOTA}\%$\uparrow$} & {\textbf{MOTA}\%$\uparrow$} & {\textbf{FP$\downarrow$}} & {\textbf{FN$\downarrow$}} & {\textbf{IDS}$\downarrow$} \\
\bottomrule % 加粗底部横线
PC3T \cite{wu20213d}  & \texttimes & 83.17 & 86.07 & 1348 & \textbf{1992} & 13 \\
PC3T \cite{wu20213d}  & \checkmark & \textbf{83.65} & \textbf{86.32} & \textbf{1252} & 2038 & \textbf{3} \\

\midrule
DFMOT \cite{wang2022deepfusionmot}  & \texttimes & 77.45 & 87.27 & \textbf{643} & 2337 & 76 \\
DFMOT \cite{wang2022deepfusionmot} & \checkmark & \textbf{77.76} & \textbf{87.29} & 724  & \textbf{2282} & \textbf{54}  \\

\bottomrule % 加粗表格底部横线
\end{tabular}%
}
\end{table}

The results clearly demonstrate that Ro\_GDIoU brings a substantial improvement to the performance of the original tracking algorithms on both the KITTI and nuScenes datasets. By integrating Ro\_GDIoU, the algorithms achieve higher accuracy in object detection and association, leading to more reliable and precise tracking, especially in complex scenarios.

\section{Conclusion}

In this work, we have developed a concise and unified 3D multi-object tracking method specifically tailored for the autonomous driving domain. Our approach has achieved SOTA performance across various datasets. Furthermore, we have standardized the perception formats of different datasets, allowing researchers to focus on the study of multi-object tracking algorithms without dealing with the cumbersome preprocessing work caused by format differences between datasets. Lastly, we have introduced a new set of evaluation metrics aimed at measuring the performance of multi-object tracking, encouraging researchers to pay attention not only to the correct matching of trajectories but also to the performance of motion attributes essential for downstream applications.
{
    \small
    \bibliographystyle{ieeenat_fullname}
    \bibliography{main}
}


\end{document}